\documentclass[journal]{IEEEtran}

\usepackage{amsmath,amsfonts}
\usepackage{algorithmic}
\usepackage{array}
\usepackage[caption=false,font=normalsize,labelfont=sf,textfont=sf]{subfig}
\usepackage{textcomp}
\usepackage{stfloats}
\usepackage{url}
\usepackage{verbatim}
\usepackage{graphicx}
\usepackage{balance}
\usepackage{cite}

\usepackage{booktabs}
\usepackage{multirow,multicol}
\usepackage{xspace}
\usepackage{dsfont}
\usepackage{pifont}
\usepackage{colortbl}
\usepackage{overpic}
\usepackage{enumitem}
\usepackage{nccmath}
\usepackage{courier}
\usepackage{array}
\usepackage{tabularray}
\usepackage{pifont}
\usepackage{microtype}
\usepackage[dvipsnames]{xcolor}

\usepackage{contour}
\usepackage{pgf,tikz}
\usepackage{tkz-euclide,tkz-graph}
\usetikzlibrary{calc}
\usepackage{pgfplots}
\pgfplotsset{
    compat=1.18,
    tick label style={font=\footnotesize},
    label style={font=\footnotesize},
    width=9cm,
    height=6cm}

\usepackage[pagebackref,breaklinks,colorlinks]{hyperref}

\begin{document}

\newcommand{\acronym}{FreeZeV2\xspace}

\newcommand{\fabiocomment}[1]{\todo[color=red!20, inline, author=Fabio]{#1}}
\newcommand{\davidecomment}[1]{\todo[color=blue!20, inline, author=Davide]{#1}}
\newcommand{\andreacomment}[1]{\todo[color=green!20, inline, author=Andrea]{#1}}
\newcommand{\warning}[1]{\textbf{\color{red!90}{#1}}}

\newcommand{\cmark}{\ding{51}}
\newcommand{\xmark}{\ding{55}}

\definecolor{myazure}{rgb}{0.8509,0.8980,0.9412}

\definecolor{freezev2darker}{RGB}{4,38,48}
\definecolor{freezev2}{RGB}{76,114,115}
\definecolor{freezev2-ms}{RGB}{53, 79, 80}
\definecolor{freeze}{RGB}{208,214,214}

\title{Accurate and efficient zero-shot 6D pose estimation\\ with frozen foundation models}

\author{
Andrea Caraffa \quad Davide Boscaini \quad Fabio Poiesi
\thanks{
\textit{(Corresponding author: Andrea Caraffa)}

Andrea Caraffa, Davide Boscaini, and Fabio Poiesi are with Fondazione Bruno Kessler (FBK), 38123 Trento, Italy.
E-mails: acaraffa@fbk.eu, dboscaini@fbk.eu, poiesi@fbk.eu.
This work was supported by the European Union’s Horizon Europe research and innovation programme under grant agreement No. 101058589 (AI-PRISM) and No. 101135707 (FORTIS).
}
}

\maketitle

\begin{abstract}
Estimating the 6D pose of objects from RGBD data is a fundamental problem in computer vision, with applications in robotics and augmented reality.
A key challenge is achieving generalization to novel objects that were not seen during training.
Most existing approaches address this by scaling up training on synthetic data tailored to the task, a process that demands substantial computational resources.
\emph{But is task-specific training really necessary for accurate and efficient 6D pose estimation of novel objects?}
To answer \emph{No!}, we introduce \acronym, the second generation of FreeZe: a training-free method that achieves strong generalization to unseen objects by leveraging geometric and vision foundation models pre-trained on unrelated data.
\acronym improves both accuracy and efficiency over FreeZe through three key contributions:
(i) a sparse feature extraction strategy that reduces inference-time computation without sacrificing accuracy;
(ii) a feature-aware scoring mechanism that improves both pose selection during RANSAC-based 3D registration and the final ranking of pose candidates; and
(iii) a modular design that supports ensembles of instance segmentation models, increasing robustness to segmentation masks errors.
We evaluate \acronym on the seven core datasets of the BOP Benchmark, where it establishes a new state-of-the-art in 6D pose estimation of unseen objects.
When using the same segmentation masks, \acronym achieves a remarkable 8× speedup over FreeZe while also improving accuracy by 5\%.
When using ensembles of segmentation models, \acronym gains an additional 8\% in accuracy while still running 2.5× faster than FreeZe.
\acronym was awarded \textit{Best Overall Method} at the BOP Challenge 2024. 
\end{abstract}

\begin{IEEEkeywords}
Object 6D pose estimation, zero-shot, training-free, foundation models.
\end{IEEEkeywords}

\section{Introduction}\label{sec:intro}

Object 6D pose estimation aims to recover the 3D position and orientation of an object within a scene from visual sensory data~\cite{hodan2018bop}.
It is a fundamental problem in computer vision, with applications in robotic manipulation~\cite{liu2023robotic, zhuang2023instance,mateo2024towards} and augmented or mixed reality~\cite{su2019deep}.
The field has progressed from traditional methods based on hand-crafted features~\cite{drost2010ppf} to modern deep learning approaches that leverage explicit supervision from task-specific data~\cite{peng2019pvnet, wang2019densefusion}.
This evolution has highlighted a critical limitation: while supervised models can achieve high accuracy on objects seen during training, their scalability is constrained by the impracticality of collecting labeled data for the vast range of objects encountered in real-world settings.

\contournumber{32}
\contourlength{.05em}

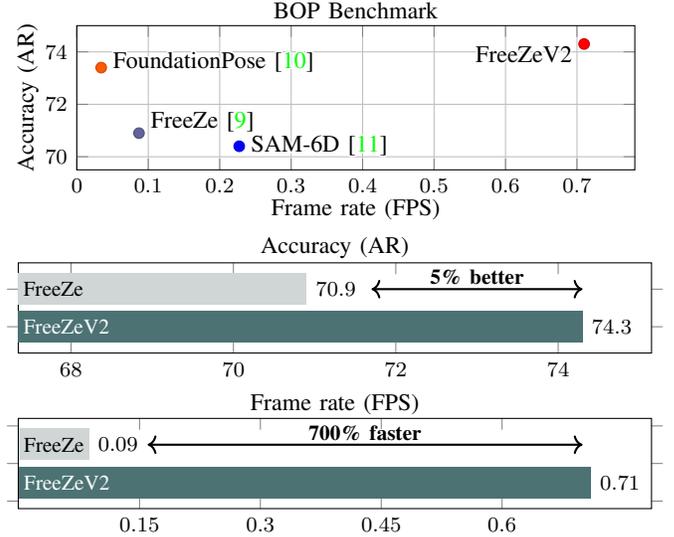
\begin{figure}[t!]
    \centering

    %
    % Comparison FPS
    %
    \begin{tikzpicture}
        \begin{axis}[
            height=3.5cm,
            title={\small BOP Benchmark},
            title style={yshift=-1.5ex}, % Adjusts vertical space between title and chart
            xlabel={\small Frame rate (FPS)},
            xlabel style={yshift=1.0ex}, % Adjusts vertical space between xlabel and chart
            ylabel={\small Accuracy (AR)},
            ylabel style={yshift=-1.0ex}, % Adjusts horizontal space between ylabel and chart
            ymajorgrids=true,
            xmajorgrids=true,
            ymin=69.5,
            ymax=75,
            xmin=0.0,
            scatter,
            only marks,
            mark size=4pt
        ]

        % FreeZeV2
        \addplot[mark options={fill=red, scale=0.5}] coordinates{(0.71, 74.3)} node[left, xshift=-1pt, yshift=-4pt] {\small \contour{white}{\acronym}};

        % FreeZe
        \addplot[mark options={scale=0.5}] coordinates{(0.087, 70.9)} node[right, xshift=1pt, yshift=4pt] {\small \contour{white}{FreeZe \cite{caraffa2024freeze}}};

        % FoundationPose
        \addplot[mark options={scale=0.5}] coordinates{(0.0341, 73.4)} node[right, xshift=1pt, yshift=2pt] {\small \contour{white}{FoundationPose \cite{wen2024foundationpose}}};

        % SAM6D
        \addplot[mark options={scale=0.5}] coordinates{(0.2273, 70.4)} node[right, xshift=1pt, yshift=0pt] {\small \contour{white}{SAM-6D \cite{lin2024sam6d}}};

        \end{axis} 
    \end{tikzpicture}

    %
    % Accuracy
    %
    \begin{tikzpicture}
        \begin{axis}[
            title={\small Accuracy (AR)},
            title style={yshift=-1.75ex}, % Adjusts vertical space between title and chart
            xbar,
            axis lines=box, % Draws borders on all sides
            xmin=68, xmax=74.5,
            width=10cm, height=4cm,
            xtick={68,70,72,74},
            ytick={1.25,1.75},
            yticklabels={,}, % This removes y-axis labels
            % xmajorgrids=true,
            % ymajorgrids=true,
            enlargelimits=0.1,
            bar width=12pt,
            nodes near coords,
            every node near coord/.append style={font=\footnotesize, color=black},
            y=1cm, % Adjusts space between bars
        ]
        % Bars
        \addplot[draw=none, fill=freeze] coordinates {(70.9,2)};
        \addplot[draw=none, fill=freezev2] coordinates {(74.3,1)};
        \node at (67.3, 1.25) [anchor=west] {\footnotesize \color{white}{\acronym}};
        \node at (67.3, 1.75) [anchor=west] {\footnotesize FreeZe};
        % Arrow
        \draw[<->, black, thick] (71.7, 1.75) -- (74.3, 1.75) node[midway, above, inner sep=1pt] {\footnotesize \contour{white}{\textbf{5\% better}}};
        \end{axis}
    \end{tikzpicture}

    \begin{tikzpicture}
        \begin{axis}[
            title={\small Frame rate (FPS)},
            title style={yshift=-1.75ex}, % Adjusts vertical space between title and chart
            xbar,
            axis lines=box, % Draws borders on all sides
            xmin=0.065, xmax=0.72,
            width=10cm, height=4cm,
            xtick={0.15,0.3,0.45,0.6},
            yticklabels={,}, % This removes y-axis labels
            % xmajorgrids=true,
            % ymajorgrids=true,
            enlargelimits=0.1,
            bar width=12pt,
            nodes near coords,
            y=1cm, % Adjusts space between bars
            every node near coord/.append style={
               /pgf/number format/fixed,
               /pgf/number format/precision=2,
               font=\footnotesize, color=black
            },
        ]
        % Bars
        \addplot[draw=none, fill=freeze] coordinates {(0.087,2)};
        \addplot[draw=none, fill=freezev2] coordinates {(0.71,1)};
        \node at (-0.005, 1.25) [anchor=west] {\footnotesize \color{white}{\acronym}};
        \node at (-0.005, 1.75) [anchor=west] {\footnotesize FreeZe};
        \draw[<->, black, thick] (0.16, 1.75) -- (0.7, 1.75) node[midway, above, inner sep=1pt] {\footnotesize \contour{white}{\textbf{700\% faster}}};
        \end{axis}
    \end{tikzpicture}

    \vspace{-3mm}
    \caption{
    (Top) Comparison of 6D pose estimation performance across various zero-shot methods using SAM-6D~\cite{lin2024sam6d} as object localization prior.
    The plot relates inference speed with average accuracy on the seven core datasets of the BOP Benchmark.
    A common trend among competitors is that higher accuracy typically comes at the expense of lower efficiency.
    Our approach breaks this trend: \acronym is both significantly more accurate and considerably faster than the other methods.
    (Bottom) \acronym introduces key architectural improvements over FreeZe~\cite{caraffa2024freeze} that result in a 5\% improvement in accuracy and a 700\% increase in efficiency, advancing toward real-time applicability.
    }
    \label{fig:teaser}

\end{figure}

As a result, increasing attention has shifted towards the generalization capabilities of 6D pose estimation methods, giving rise to a taxonomy that classifies them as \emph{instance-level}, \emph{category-level}, or \emph{zero-shot}.
Instance-level methods~\cite{peng2019pvnet, wang2019densefusion} require training on the exact object instances encountered at test time and lack the ability to generalize to unseen objects.
Category-level methods~\cite{di2024zero123, li2025gce, wang2024gs} generalize across different instances within the same object category, but are limited to categories seen during training.
Both settings operate under the closed-set assumption, which limits their real-world applicability.
In contrast, zero-shot methods adopt the open-set assumption, addressing 6D pose estimation of novel objects without relying on explicit supervision on predefined instances or categories.
Zero-shot methods can be further classified as \emph{training-based} or \emph{training-free}.
Training-based methods~\cite{chen2024zeropose, moon2024genflow, huang2024matchu, wen2024foundationpose, lin2024sam6d, labbe2022megapose, nguyen2024gigapose} aim to generalize by training on large-scale synthetic 6D pose datasets~\cite{wen2024foundationpose,labbe2022megapose,hamza2025distilling}, but face several limitations: they are sensitive to the quality and diversity of synthetic data, require substantial computational resources and long training times, and are affected by the synthetic-to-real domain gap.
Instead, training-free methods~\cite{caraffa2024freeze, ornek2024foundpose, ausserlechner2024zs6d} sidestep these challenges by leveraging the general-purpose knowledge embedded in foundation models.

Alternatively, 6D pose estimation methods can be classified as either \emph{model-based} or \emph{model-free}, depending on the information available about the object at inference time.
Model-based approaches~\cite{caraffa2024freeze, wen2024foundationpose, lin2024sam6d} assume access to the 3D model of the object, while model-free methods~\cite{sun2022onepose, he2022onepose++, liu2022gen6d, liu2025novel, di2025instantpose, liu2024unopose, liu2025one2any,corsetti2024open,corsetti2024high} operate with only a few images capturing the object from multiple viewpoints.

Our previous work, FreeZe~\cite{caraffa2024freeze}, operates in the model-based, training-free, zero-shot setting.
It estimates 6D poses by combining \emph{feature extraction} and \emph{feature matching}:
FreeZe fuses visual and geometric features from frozen foundation models pre-trained on task-unrelated data, establishes object-to-scene correspondences, and estimates the object 6D pose via RANSAC-based registration.
This design enables strong generalization to unseen objects without requiring any task-specific training.
However, FreeZe's reliance on dense feature extraction and registration incurs substantial computational overhead, limiting its applicability in real-world scenarios.

In this paper, we introduce \acronym, the second generation of FreeZe, which addresses these limitations through a simple yet effective principle: retain dense and expressive features for the object's 3D model, while extracting only sparse features from the RGBD scene.
Object features are precomputed offline, whereas scene features are extracted online only at a sparse set of informative locations within the region identified by the segmentation mask.
Candidate poses are estimated via sparse-to-dense feature matching using RANSAC-based registration, refined using Iterative Closest Point (ICP)~\cite{arun1987least}, and evaluated using a novel scoring mechanism that promotes both geometric alignment and feature similarity.
The scoring function also plays a key role in handling segmentation uncertainty, enabling robust pose selection across multiple segmentation mask candidates.
This results in a modular and efficient pipeline that delivers high accuracy at a fraction of the computational cost.
Moreover, the same architecture extends naturally to both \emph{6D localization}, where the number and identity of object instances in the scene is known, and \emph{6D detection}, where this information is withheld.

We evaluate \acronym on the BOP Benchmark~\cite{hodan2018bop}, the most comprehensive benchmark for object 6D pose estimation, covering seven RGBD datasets with a wide range of object types, occlusion levels, textures, and clutter.
\acronym establishes a new state-of-the-art in both 6D localization and 6D detection, outperforming the most prominent zero-shot methods by a large margin in both accuracy and efficiency (Fig.~\ref{fig:teaser}, top).
Specifically, \acronym achieves an 8× speed-up and a +3.4\,AR improvement over FreeZe (Fig.~\ref{fig:teaser}, bottom).
\acronym was awarded \textit{Best Overall Method} at the BOP Challenge 2024 for both 6D localization and 6D detection of unseen objects~\cite{nguyen2025bop}.

In summary, our main contributions are:
\begin{itemize}[noitemsep,nolistsep,leftmargin=*]
\item We propose a sparse-to-dense feature matching strategy that improves efficiency without sacrificing accuracy.
\item We introduce a scoring mechanism that jointly considers geometric alignment and feature similarity to improve both pose hypothesis selection during RANSAC-based 3D registration and the final ranking across pose candidates.
\item We design a modular pipeline that supports ensembles of zero-shot instance segmentation models, increasing robustness to segmentation masks errors.
\item We extend our method to the more challenging 6D detection setting, where the number and identity of object instances present in the scene are unknown at inference time.
\end{itemize}

\section{Related works}\label{sec:related}

\begin{table}[t]
    \renewcommand{\arraystretch}{0.9}
    \tabcolsep 4pt
    \caption{
    Comparison of 6D pose estimation methods by input data type (D: depth, RGB: color, RGBD: both), need for task-specific training, use of foundation models (FM), core strategy (template vs feature matching), and feature type (V: vision, G: geometric).
    }
    \label{tab:related}
    \vspace{-2mm}
    \resizebox{\columnwidth}{!}{%
    \begin{tabular}{lccccc}
        \toprule
        \multirow{2}{*}{Method} & \multirow{2}{*}{Input} & Training & \multirow{2}{*}{FM} & Template & Feature \\
        & & free & & matching & matching \\
        \toprule
        OVE6D~\cite{cai2022ove6d} & D & & & G &\\
        GCPose~\cite{zhao2023learning} & D & & & & G\\
        Drost et al.~\cite{drost2010ppf} & D & \cmark & & & G\\
        \midrule
        MegaPose~\cite{labbe2022megapose} & RGB & & & V &\\
        Nguyen et al.~\cite{nguyen2022templates} & RGB & & & V &\\
        Wang et al.~\cite{wang2024object}& RGB & &  \cmark & V &\\
        GigaPose~\cite{nguyen2024gigapose} & RGB & & & V & V\\
        OSOP~\cite{shugurov2022osop} & RGB & & & V & V\\
        ZS6D~\cite{ausserlechner2024zs6d} & RGB & \cmark & \cmark & V & V\\
        FoundPose~\cite{ornek2024foundpose} & RGB & \cmark & \cmark & V & V\\
        DZOP~\cite{von2024diffusion} & RGB & \cmark & \cmark & V & V\\
        \midrule
        ZePHyR~\cite{okorn2021zephyr} & RGBD &  & & V+G\\
        GenFlow~\cite{moon2024genflow} & RGBD & &  & V+G\\
        FoundationPose~\cite{wen2024foundationpose} & RGBD & & & V+G &\\
        ZeroPose~\cite{chen2024zeropose} & RGBD & & \cmark & V & G \\
        SAM-6D~\cite{lin2024sam6d} & RGBD & & & & V+G\\
        MatchU~\cite{huang2024matchu} & RGBD & & & & V+G\\
        FreeZeV2 (ours) & RGBD & \cmark & \cmark & & V+G\\
        \bottomrule
    \end{tabular}
    }
\end{table}

\subsection{6D pose estimation of unseen objects}

Recent research in 6D pose estimation increasingly focuses on the zero-shot setting, where objects encountered at inference time are not seen during training~\cite{liu2024deep}.
Tab.~\ref{tab:related} summarizes representative zero-shot 6D pose estimation methods by input type, training requirements, use of foundation models, core strategy (template vs feature matching), and feature modality (vision vs geometric).
\emph{Training-based} methods can achieve generalization to unseen objects via large-scale training on synthetic data~\cite{labbe2022megapose, wen2024foundationpose}, while 
\emph{training-free} methods rely on hand-crafted features~\cite{drost2010ppf} or foundation models pre-trained on unrelated web-scale data~\cite{ausserlechner2024zs6d, ornek2024foundpose, caraffa2024freeze}.
Following~\cite{liu2024deep}, methods are further categorized into \emph{template matching} and \emph{feature matching}.
Template matching estimates object's pose by comparing the input scene image to rendered templates of the object's 3D model from various viewpoints, selecting the best match, typically followed by refinement~\cite{okorn2021zephyr, cai2022ove6d, wang2024object, moon2024genflow}.
Feature matching estimates object's pose by establishing 2D-3D (or 3D-3D) correspondences between local features extracted from the input scene image and the object's 3D model, followed by PnP (or 3D) registration~\cite{drost2010ppf, zhao2023learning, lin2024sam6d, huang2024matchu}.
Some template matching approaches first retrieve the best-matching template and then establish 2D-3D correspondences against it for PnP-based pose estimation~\cite{shugurov2022osop, nguyen2024gigapose, ausserlechner2024zs6d, ornek2024foundpose, von2024diffusion}.
Below, we provide a detailed comparison of their differences.

\vspace{1mm}
\noindent\emph{Training-based template matching methods}
improve generalization by generating large-scale synthetic  datasets using rendering engines \cite{labbe2022megapose,shugurov2022osop,nguyen2024gigapose}.
For example, MegaPose~\cite{labbe2022megapose} renders 2 million images of over 20 thousand objects from ShapeNet~\cite{chang2015shapenet} and GSO~\cite{downs2022google}, while FoundationPose~\cite{wen2024foundationpose} uses LLM-aided texture augmentation to scale up data and generate 1.2 million images with greater 3D model diversity.
In contrast, we demonstrate that such large-scale, task-specific 6D pose datasets are not necessary for accurate pose estimation.
OSOP~\cite{shugurov2022osop} and GigaPose~\cite{nguyen2024gigapose} perform template matching using visual features learned with convolutional neural network (CNN) or Vision Transformer (ViT) backbones~\cite{nguyen2022templates}, whereas Wang et al.~\cite{wang2024object} rely on frozen diffusion features combined with an aggregation module trained with 6D pose supervision.
Unlike training-based template matching approaches, our method performs feature matching using pre-trained visual and geometric features, requiring no task-specific training or synthetic data, and still achieves strong generalization to unseen objects.

\vspace{1mm}
\noindent\emph{Training-based feature matching methods}
learn local features using either visual~\cite{dosovitskiy2021vit} or geometric encoders~\cite{thomas2019kpconv, qin2022geotransformer, yu2023rotation}.
GCPose~\cite{zhao2023learning} trains geometric features on depth images for 3D registration.
ZeroPose~\cite{chen2024zeropose} uses frozen visual features for template retrieval, followed by feature matching with learned geometric descriptors.
SAM-6D~\cite{lin2024sam6d} and MatchU~\cite{huang2024matchu}
train custom networks to fuse visual and geometric features.
In contrast, our method leverages frozen vision and geometric foundation models, benefiting from their strong generalization capabilities without requiring any task-specific training.

\vspace{1mm}
\noindent\emph{Training-free template-matching methods}
provide an alternative to synthetic data-driven approaches by leveraging frozen foundation models for feature extraction without any finetuning.
ZS6D~\cite{ausserlechner2024zs6d}, FoundPose~\cite{ornek2024foundpose}, and DZOP~\cite{wang2024object} match target images to rendered templates using frozen vision transformers, establishing 2D-3D correspondences for PnP-based pose estimation.
However, these approaches rely solely on visual features and often fail on textureless objects or under challenging lighting conditions~\cite{hodan2017tless,tudl}.
To address these limitations, we propose the first training-free method that leverages the synergy between vision and geometric foundation models for 3D-3D matching, improving robustness to difficult scenarios.

\vspace{1mm}
\noindent\emph{Training-free feature-matching methods} traditionally rely on hand-crafted point-pair features~\cite{drost2010ppf} that operate only on depth data.
While effective in certain cases, these methods struggle with geometric symmetries and disregard the rich semantic cues present in RGB images.
In contrast, our approach fuses GeDi rotation-invariant geometric descriptors~\cite{poiesi2023gedi} with DINOv2 semantically rich visual features~\cite{oquab2023dinov2}, resolving symmetry-induced pose ambiguities and delivering more reliable pose estimates across diverse objects and lighting conditions.

\subsection{Vision foundation models}
Vision foundation models produce general-purpose image features that enable zero-shot generalization across diverse tasks without finetuning. CLIP~\cite{radford2021clip} is trained contrastively on massive image–text pairs, aligning visual concepts with language in a shared embedding space.
While CLIP excels at global semantic retrieval, its embeddings lack the spatial precision needed for accurate 6D pose estimation.
By contrast, DINO~\cite{caron2021dinov1} learns patch-level representations through self-supervised distillation, yielding both spatially consistent and semantically rich features. DINOv2~\cite{oquab2023dinov2} further scales model size and data to improve pre-training stability and efficiency. In this work, we use a frozen DINOv2 encoder to extract spatially precise and semantically rich features, enabling sparse-to-dense correspondence matching for accurate 6D pose estimation.

\subsection{Geometric feature extractors}
Geometric feature extractors project 3D points into high-dimensional representations that capture local geometric structures.
These methods broadly operate at either the voxel or point level.
Voxel-based approaches leverage sparse 3D convolutions to efficiently process non-empty voxels~\cite{choy20194d, choy2019fcgf, corsetti2023fcgf6d}.
Point-level methods operate directly on raw point clouds.
Permutation-invariant MLPs encode points independently and aggregate local information through pooling~\cite{qi2017pn, qi2017pn2, qian2022pointnext}.
Point convolutional methods learn local geometric patterns using local charting mechanisms~\cite{masci2015gcnn, MoNet, SplineCNN, thomas2019kpconv}.
Graph-based~\cite{wang2019dgcnn} and Transformer-based~\cite{zhao2021ptv1, wu2022ptv2, wu2024ptv3} approaches introduce large-range interactions via message passing and attention mechanisms, respectively.
More recently, mixer-based approaches~\cite{ma2022pointmlp, choe2022pointmixer, boscaini2023patchmixer} aggregate point-level features with token-mixing MLPs~\cite{tolstikhin2021mlpmixer, trockman2022convmixer} as a more efficient alternative to self-attention~\cite{vaswani2017attention, dosovitskiy2021vit}.
In this work, we use a frozen GeDi~\cite{poiesi2023gedi} encoder which ensures rotation-invariant feature extraction by aligning local reference frames and demonstrates strong generalization across datasets.
\section{Our method}\label{sec:method}

\begin{figure*}[t!]
    \centering
    \begin{overpic}[trim=17 0 68 0, clip, width=\textwidth]{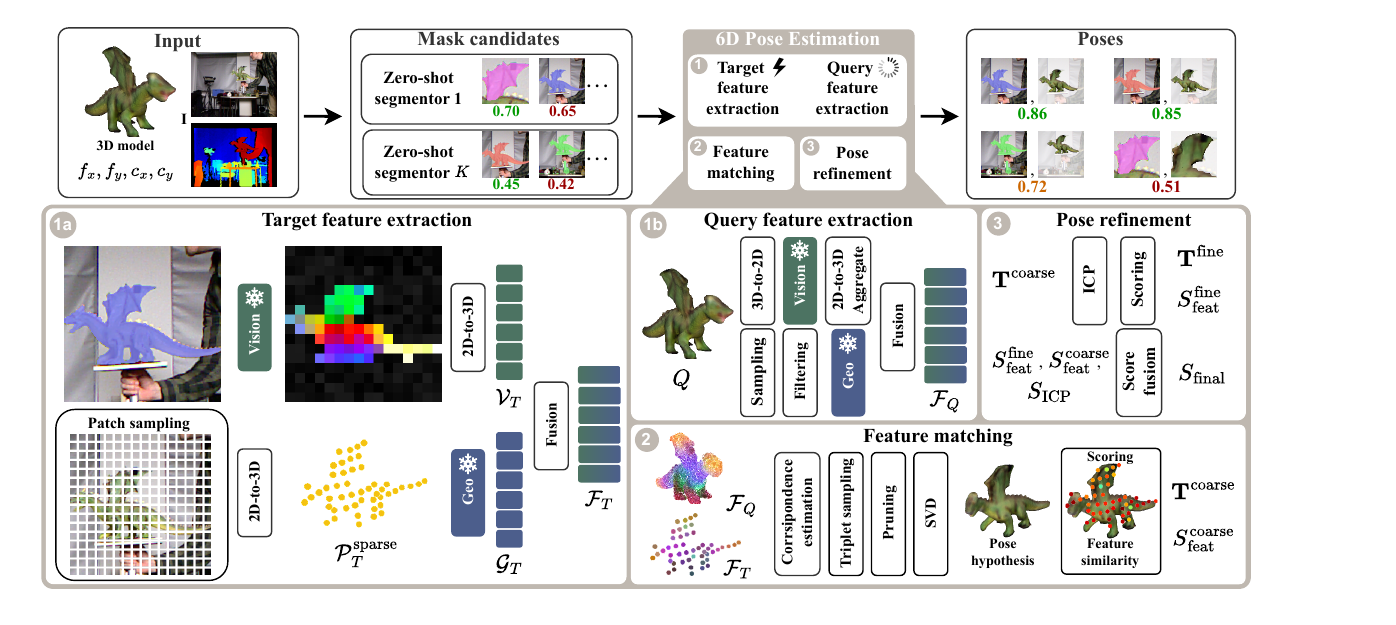}
    \end{overpic}
    \vspace{-10mm}
\caption{
\textbf{Overview of the proposed \acronym pipeline.}
Given a query object $Q$ represented as a 3D model and an RGBD test image $\mathbf{I}$, our method estimates the 6D pose of $Q$ through three stages: (1) \textit{feature extraction}, (2) \textit{feature matching}, and (3) \textit{pose refinement}.  
\textit{Top row:} Candidate masks are generated by applying multiple zero-shot instance segmentation models independently. All masks are retained and processed in parallel.
\textit{Stage 1 (Feature extraction):}  
(1a) \textit{Target features} are extracted online.
A $16 \times 16$ grid of 2D patches is sampled from each candidate mask and backprojected into 3D to obtain a sparse point cloud $\mathcal{P}_T^\text{sparse}$.
Visual features are computed using a frozen DINOv2 encoder and fused with geometric descriptors from a frozen GeDi encoder, resulting in fused target features $\mathcal{F}^T$.  
(1b) \textit{Query features} $\mathcal{F}^Q$ are precomputed offline.
Dense 3D points $\mathcal{P}_Q$ are sampled on the model surface, and each point is encoded with fused geometric and visual features aggregated across multiple rendered views.  
\textit{Stage 2 (Feature matching):}  
Feature correspondences are established via nearest-neighbor search in the fused feature space. RANSAC generates pose hypotheses $\mathbf{T}^\text{coarse}$, each scored using a feature-aware metric $S_\text{feat}$ based on feature similarity.  
\textit{Stage 3 (Refinement):}  
The top-scoring coarse pose is refined via ICP to obtain $\mathbf{T}^\text{fine}$.
A final score $S_\text{final}$ combines feature similarity at coarse and fine levels, and ICP inlier ratio.
Duplicate hypotheses are removed through non-maximum suppression to produce the final pose estimates.
}
\label{fig:diagram}
\end{figure*}

\subsection{Problem formulation}
Given the 3D model of a \emph{query object} $Q$ and an RGBD image $\mathbf{I}$ capturing a scene, we refer to the visible instance of $Q$ within $\mathbf{I}$ as the \emph{target object} $T$.
The goal of 6D pose estimation is to recover the six degrees-of-freedom (DoF) transformation $\mathbf{T} = (\mathbf{R}, \mathbf{t}) \in SE(3)$ that aligns $Q$ to $T$, where $\mathbf{R} \in SO(3)$ denotes the 3DoF rotation matrix and $\mathbf{t} \in \mathbb{R}^3$ is the 3DoF translation vector.
Estimating $\mathbf{T}$ in real-world scenarios is challenging due to significant discrepancies between $Q$ and $T$.
While $Q$ is represented as a complete 3D model with uniformly sampled surface points, $T$ is captured by an RGBD sensor, resulting in a partial and noisy point cloud with uneven point density.
Moreover, $T$ may be occluded by surrounding objects, and its photometric appearance can vary significantly due to lighting changes or shadows cast by the environment.

The 6D pose estimation problem can be evaluated under two settings, namely \emph{6D localization} and \emph{6D detection}, which differ in the assumptions made at inference time.
The 6D localization setting assumes prior knowledge about which objects are visible and how many instances of them are present in the input image.
In contrast, the 6D detection setting assumes no prior knowledge about which objects appear in the scene or how many instances are present.
These evaluation protocols follow the definitions outlined by the BOP Benchmark~\cite{nguyen2025bop}.

\subsection{Overview}

\acronym consists of three main stages: (1) \textit{feature extraction}, (2) \textit{feature matching}, and (3) \textit{pose refinement}, as illustrated in Fig.~\ref{fig:diagram}.
Before these stages, we perform zero-shot object segmentation to localize $T$ within $\mathbf{I}$, generating a set of candidate segmentation masks.
For each candidate mask, we estimate a 6D pose and compute a confidence score, which is then used to rank the resulting 6D pose candidates.

\noindent\textbf{(Pre-processing) Segmentation.}
During preprocessing, we apply one or more zero-shot instance segmentation models to the input image $\mathbf{I}$ to extract candidate segmentation masks.
The most confident masks are retained and processed independently in subsequent stages.
Using an ensemble of segmentation models increases robustness to segmentation errors.

\noindent\textbf{(1) Feature extraction.}
We extract visual and geometric features with frozen foundation models for both query and target objects.
Query features are precomputed offline by densely sampling points on the query 3D model and encoding them with a combination of geometric and visual (from rendered views) features.
Target features are extracted online.
We sample $16 \times 16$ patch centers, compute 2D visual features, 3D-lift them using depth, and finally extract geometric features for 3D-lifted patch centers from local 3D neighborhoods.

\noindent\textbf{(2) Feature matching.}
We perform \emph{sparse-to-dense feature matching} between sparse target features and dense query features by searching for nearest neighbors in the fused feature space.
The resulting 3D-3D correspondences are used to generate multiple 6D pose hypotheses via RANSAC-based 3D registration. 
Each pose hypothesis is then scored using the proposed \emph{feature-aware similarity} metric, which jointly assesses geometric alignment and semantic consistency.

\noindent\textbf{(3) Pose refinement.}
The best pose hypothesis is refined using ICP.
We then compute a pose confidence score by assessing feature similarity at both coarse and fine levels, together with the ICP inlier ratio to quantify geometric alignment.

\noindent\textbf{Overall pipeline.}
Steps (1)--(3) are repeated for each candidate mask, resulting in as many candidate poses.
We apply Non-Maximum Suppression (NMS) across the translation component of candidate poses to eliminate duplicates.
The following sections provide a detailed description of each component.

\subsection{Segmentation}

Given the input image $\mathbf{I}$, we apply $K$ off-the-shelf, model-based, zero-shot instance segmentation methods $\{ \mathcal{S}_k \mid k=1, \dots, K \}$ to generate a set of \emph{candidate masks} for the target object $T$~\cite{nguyen2023cnos, lin2024sam6d, lu2024adapting}.
Each method $\mathcal{S}_k$ first produces object-agnostic segmentation masks using promptable segmentation models such as SAM~\cite{kirillov2023sam} or FastSAM~\cite{zhao2023fastsam}, and then assigns a confidence score to each mask based on its similarity to rendered templates of the query object's 3D model.
We incorporate multiple methods as they differ in how these confidence scores are computed.
For each method $\mathcal{S}_k$, we retain the top $M$ masks with the highest scores.
In the 6D localization setting, $M$ scales with the number of target object instances $N$.
In our experiments, we chose $M > N$ to account for potential calibration errors, i.e., cases where $\mathcal{S}_k$ assigns higher confidence to inaccurate masks than to accurate ones.
In the 6D detection setting, $M$ is a constant, and masks with confidence below $\tau_\text{mask}$ are discarded.
The final set of candidate masks is the union of all retained masks across methods.
We do not merge or filter overlapping masks to avoid discarding potentially useful detections.
Each candidate mask is then processed independently by the subsequent feature extraction, matching, and refinement modules.

\subsection{Feature Extraction}

We extract visual and geometric features from $Q$ and $T$ using frozen foundation models pre-trained on data unrelated to 6D pose estimation.
For the visual modality, we use DINOv2~\cite{oquab2023dinov2}, a vision foundation model trained on web-scale 2D images using self-supervised learning.
For the geometric modality, we use GeDi~\cite{poiesi2023gedi}, a geometric encoder trained on 3D point clouds from indoor scenes using supervised contrastive learning.
These models are selected for their strong generalization capabilities: DINOv2 produces semantically rich, general-purpose features that transfer well across diverse visual tasks, while GeDi provides highly distinctive, rotation-invariant descriptors that generalize effectively across diverse 3D data.
Given the differing nature of $Q$ (a 3D model) and $T$ (an RGBD image), we apply custom feature extraction strategies for each of them, as detailed below.

\noindent\textbf{Query feature extraction.}  
We begin by sampling $N_Q^\text{raw}$ surface points from the 3D model using Poisson disk sampling~\cite{bridson2007fast}, which ensures uniform coverage with a minimum inter-point distance.
This produces the query point cloud $\mathcal{P}_Q^\text{raw} = \{ \mathbf{p}_Q^i \in \mathbb{R}^3 \mid i=1, \dots, N_Q^\text{raw} \}$.  
To extract visual features, we render the model from multiple viewpoints and process each view with the frozen DINOv2 encoder.  
Patch features are upsampled to pixel resolution via bilinear interpolation and backprojected onto the 3D surface using known camera intrinsics and rendered depth.  
For each surface point, features from all viewpoints where it is visible are aggregated to form a unified visual feature.

To ensure feature reliability, we retain only points visible in at least $V$ distinct views, resulting in the final query point cloud $\mathcal{P}_Q = \{ \mathbf{p}_Q^i \in \mathbb{R}^3 \mid i=1, \dots, N_Q \} \subseteq \mathcal{P}_Q^\text{raw}$.
For each $\mathbf{p}_Q^i \in \mathcal{P}_Q$, we compute a visual descriptor $\mathbf{f}_\text{vis}^i \in \mathbb{R}^{D_\text{vis}}$ by weighted average of the per-view DINOv2 features, and a geometric descriptor $\mathbf{f}_\text{geo}^i \in \mathbb{R}^{D_\text{geo}}$ using the frozen GeDi encoder.

The final fused descriptor $\mathbf{f}_Q^i \in \mathbb{R}^D$ is defined as:
\begin{equation}\label{eq:query_fused_descriptor}
    \mathbf{f}_Q^i = \left[ \operatorname{norm}(\text{PCA}(\mathbf{f}_{\text{vis}}^i))\,,\, \operatorname{norm}(\mathbf{f}_{\text{geo}}^i) \right],
\end{equation}
where $[\cdot\,,\,\cdot]$ denotes concatenation, $\operatorname{norm}(\cdot)$ represents $L_2$ normalization, and the Principal Component Analysis (PCA) operator~\cite{jolliffe2002principal} reduces the dimensionality of visual descriptors to match that of the geometric features, ensuring a balanced contribution from both modalities.

The final query representation is the fused feature point cloud $\mathcal{F}_Q = \{ (\mathbf{p}_Q^i, \mathbf{f}_Q^i) \mid i=1, \dots, N_Q \}$, where $\mathbf{f}_Q^i$ has dimensionality $2D_{\text{geo}}$.
This representation is precomputed once per object and reused across all test images during inference.

\noindent\textbf{Target feature extraction.}  
For each candidate mask, we begin by sampling a $16 \times 16$ grid of patches within the smallest axis-aligned square bounding box enclosing the mask.  
Let $\{ \mathbf{u}_T^j \in \mathbb{R}^2 \mid j=1, \dots, N_T \}$ denote the 2D centers of those patches that fall inside the mask, with $N_T \le 256$ by design.
We extract visual features $\mathbf{f}_\text{vis}^j \in \mathbb{R}^{D_\text{vis}}$ at these locations using the frozen DINOv2 encoder.  
The 2D coordinates $\mathbf{u}_T^j$ are backprojected into 3D using the depth map and known camera intrinsics, resulting in a sparse target point cloud $\mathcal{P}_T^\text{sparse} = \{ \mathbf{p}_T^j \in \mathbb{R}^3 \mid j = 1, \dots, N_T \}$.  
Unlike in the query branch, no per-pixel interpolation is required, as features are directly assigned to the backprojected points $\mathbf{p}_T^j$.
This produces a sparse but informative set of points enriched with visual semantic context.  

To compute geometric features, we first extract a denser point cloud $\mathcal{P}_T^\text{dense} \subset \mathbb{R}^{N_T^\text{dense} \times 3}$ from the depth map within the same masked region.  
For each $\mathbf{p}_T^j \in \mathcal{P}_T^\text{sparse}$, we compute a geometric descriptor $\mathbf{f}_\text{geo}^j \in \mathbb{R}^{D_\text{geo}}$ using the frozen GeDi encoder, based on its local neighborhood in $\mathcal{P}_T^\text{dense}$.

Each sparse target point is then associated with a fused descriptor $\mathbf{f}_T^j \in \mathbb{R}^{D}$, defined as:
\begin{equation}\label{eq:target_fused_descriptor}
    \mathbf{f}_T^j = \left[ \operatorname{norm}(\text{PCA}(\mathbf{f}_{\text{vis}}^j))\,,\, \operatorname{norm}(\mathbf{f}_{\text{geo}}^j) \right],
\end{equation}
where $[\cdot\,,\,\cdot]$ denotes concatenation, $\operatorname{norm}(\cdot)$ is $L_2$ normalization, and PCA is applied to reduce visual feature dimensionality, matching the representation used in the query branch.

The final target representation is the fused feature point cloud $\mathcal{F}_T = \{ (\mathbf{p}_T^j, \mathbf{f}_T^j) \mid j=1, \dots, N_T \}$, where $\mathbf{f}_T^j$ has dimensionality $2D_{\text{geo}}$.
In contrast to the query, $\mathcal{F}_T$ is computed online at test time, as it depends on the input image.

\subsection{Feature Matching}
Given the fused features $\mathcal{F}_Q$ of the query object $Q$ and the fused features $\mathcal{F}_T$ of the target object $T$, we establish 3D-3D correspondences and generate coarse pose hypotheses via RANSAC-based registration.  
We perform sparse-to-dense feature matching from the sparse target point cloud to the dense query model in the fused feature space.

\noindent\textbf{Correspondence estimation.}  
For each target point $\mathbf{p}_T^j \in \mathcal{P}_T^\text{sparse}$, we identify its $k$ nearest neighbors in the query point cloud $\mathcal{P}_Q$ based on cosine similarity between fused features.  
The resulting set of correspondences is defined as:
\begin{equation}\label{eq:corres_set}
    \mathcal{C} = \left\{ \left( \mathbf{p}_T^j, \mathbf{p}_Q^i \right) \,\middle|\, i \in \operatorname{TopK} \left( \arg\max_{i'} \cos(\mathbf{f}_T^j, \mathbf{f}_Q^{i'}) \right) \right\},
\end{equation}
where $\cos(\mathbf{f}_T^j, \mathbf{f}_Q^{i'}) = ( \mathbf{f}_T^j \cdot \mathbf{f}_Q^{i'} ) / \|\mathbf{f}_T^j\|_2 \|\mathbf{f}_Q^{i'}\|_2 $ is the cosine similarity and the operator $\operatorname{TopK}(\cdot)$ selects the $k$ query points with the highest cosine similarity to each target point, increasing robustness to feature noise and ambiguity.

\noindent\textbf{Pose hypothesis generation.}  
We apply RANSAC to robustly estimate a coarse 6D transformation $\mathbf{T}^\text{coarse} = (\mathbf{R}, \mathbf{t}) \in SE(3)$ from the set of correspondences $\mathcal{C}$.  
At each iteration, RANSAC samples a triplet of correspondences and estimates the rigid transformation using singular value decomposition (SVD)~\cite{trefethen2022numerical}.  
To reduce the impact of spurious matches and improve efficiency, we include a fast pruning step that filters correspondences before transformation estimation.  
Specifically, we reject triplets in which the distances between matched points exceed a geometric threshold or where relative edge lengths between query and target pairs are inconsistent.  
Only valid transformations are used to compute the pose hypothesis.

\noindent\textbf{Hypothesis scoring.}  
Each pose hypothesis is evaluated using a feature-aware similarity score that accounts for both geometric alignment and semantic consistency in feature space.  
For each hypothesis, we define the inlier set $\mathcal{I}$ as the set of correspondences for which the transformed query points lie within a predefined distance threshold $\tau_\text{inlier}$ from their corresponding target points:
\begin{equation}\label{eq:ransac_inliers}
\left\| \mathbf{R} \mathbf{p}_Q^i + \mathbf{t} - \mathbf{p}_T^j \right\| < \tau_\text{inlier}.
\end{equation}
Given $\mathcal{I}$, we define the coarse-level score as:
\begin{equation}\label{eq:score_feat}
S_\text{feat}^\text{coarse} = \frac{1}{|\mathcal{P}_T^\text{sparse}|} \sum_{(j,i) \in \mathcal{I}} \cos \left( \mathbf{f}_T^j, \mathbf{f}_Q^i \right),
\end{equation}
where $\cos(\cdot, \cdot)$ is the cosine similarity between $L_2$-normalized fused features and $\lvert \cdot \rvert$ denotes cardinality.  
This score promotes pose hypotheses that align not only geometrically but also semantically, leading to improved robustness in the presence of challenging scenarios involving occlusions, clutter, or suboptimal masks, where purely geometric alignment may fail.
The pose hypothesis with the highest score $S_\text{feat}^\text{coarse}$ is selected as the coarse pose estimate.

\subsection{Pose Refinement}

For each candidate mask, we refine the coarse pose associated with the highest confidence score via geometric alignment and compute a final confidence score to robustly rank the refined poses across all candidate masks.

\noindent\textbf{ICP refinement.}
The coarse pose $\mathbf{T}^\text{coarse}$ is refined via ICP~\cite{besl1992icp} to obtain a more accurate estimate $\mathbf{T}^\text{fine}$.  
This refinement aligns $\mathcal{P}_Q$ to  $\mathcal{P}_T^\text{dense}$ by maximizing the inlier ratio:
\begin{equation}\label{eq:s_icp}
    S_\text{ICP} = \frac{| \{ (\textbf{p}_Q^i, \textbf{p}_T^j) \; \text{s.t.} \; \lVert \textbf{T}^\text{fine} \textbf{p}_Q^i - \textbf{p}_T^j \rVert < \tau_\text{ICP} \} |}{|\mathcal{P}_Q|},
\end{equation}
where $\textbf{p}_Q^i \in \mathcal{P}_Q$, $\textbf{p}_T^j \in \mathcal{P}_T^\text{dense}$, and $\tau_\text{ICP}$ is a predefined distance threshold.
We also recompute the feature-based similarity score at the refined pose, denoted by $S_\text{feat}^\text{fine}$, using the same formulation provided in Eq.~\eqref{eq:score_feat}.

\noindent\textbf{Final scoring.}
To rank the ICP-refined poses, we combine geometric alignment and feature similarity at both coarse and fine levels into a single score:
\begin{equation}\label{eq:final_score}
    S_\text{final} = (S_\text{feat}^\text{coarse})^\alpha \cdot (S_\text{feat}^\text{fine})^\beta \cdot (S_\text{ICP})^\gamma,
\end{equation}
where $\alpha$, $\beta$, and $\gamma$ are hyper-parameters that weight the contribution of each term.
Note that $S_\text{final}$ is independent of the initial confidence assigned to the segmentation masks.
This allows accurate poses to emerge even from low-confidence masks: for example, a segmentation mask with low confidence may still yield an accurate pose if it exhibits strong feature similarity and geometric alignment.
By decoupling pose quality from segmentation confidence, this scoring mechanism enhances robustness to segmentation errors, allowing the pipeline to recover accurate predictions that might otherwise be discarded.

\noindent\textbf{Non-Maximum Suppression.} Lastly, we apply NMS to the refined poses to eliminate duplicates based on their translation distance and retain only distinct object instance poses.

\section{Results}\label{sec:results}

\subsection{Experimental setup}\label{sec:setup}

We evaluate our method using segmentation masks generated by four zero-shot models: CNOS~\cite{nguyen2023cnos}, SAM-6D~\cite{lin2024sam6d}, NIDS~\cite{lu2024adapting}, and MUSE~\cite{muse_bop2024}. These models are used either individually or as an ensemble.
We render 162 templates per object using the camera viewpoints proposed by CNOS~\cite{nguyen2023cnos} and we filter $\mathcal{P}_Q^{\text{raw}}$ by keeping only the points that are visible from at least $V=18$ views, thus creating $\mathcal{P}_Q$.
To ensure consistency, all 3D models are rescaled such that the object occupies approximately 50\% of the width and height in a $480 \times 480$ image.
We compute visual features using ViT-giant DINOv2~\cite{oquab2023dinov2}, extracting patch features from intermediate layers as proposed in FoundPose~\cite{ornek2024foundpose}. 
For $\mathcal{P}_Q$ and $\mathcal{P}_T^{\text{dense}}$ we use 5k and 3k points, respectively, while $\mathcal{P}_T^{\text{sparse}}$ includes up to 256 points.
For correspondence estimation, we use a top-$k$ strategy with $k = 10$.
In the 6D localization setting, we set the number of masks per segmentation model to $M = N + 1$, where $N$ is the number of expected object instances.  
In the 6D detection setting, we set $M = 100$ and discard masks with confidence scores below a threshold $\tau_\text{mask} = 0.4$.
We compute geometric features using GeDi \cite{poiesi2023gedi} at two scales.
We extract 32-dimensional GeDi features from local neighbours occupying
30\% and 40\% of the object's diameter.
After PCA and fusion with visual features, the final fused feature dimension is 128.
Thresholds $\tau_\text{inlier}$  and $\tau_\text{ICP}$ are set to 3\% of the object's diameter.
We run parallelized RANSAC on GPU for 10,000 iterations.
We conduct experiments using a NVIDIA Tesla A40 GPU and an Intel(R) Xeon(R) Silver 4316 CPU operating at 2.30GHz.

\subsection{Datasets}\label{sec:datasets}

We evaluate \acronym on the seven core datasets of the BOP Benchmark, namely LM-O~\cite{brachmann2014learning}, T-LESS~\cite{hodan2017tless}, TUD-L~\cite{tudl}, IC-BIN~\cite{doumanoglou2016recovering}, ITODD~\cite{itodd}, HB~\cite{hb}, and YCB-V~\cite{xiang2018posecnn}, which cover a wide range of object types and scene conditions.
Each dataset provides RGBD images, 3D object models, and the intrinsic parameters of the acquisition sensor.  

\vspace{1mm}
\noindent\textbf{Object geometry.}  
T-LESS and ITODD contain industrial objects such as electrical and mechanical parts, characterized by planar surfaces, sharp edges, and hollow structures. Their 3D models are synthetic CAD meshes.  
The remaining datasets (LM-O, TUD-L, IC-BIN, HB, and YCB-V) include everyday items such as food containers, toys, and tools, typically with smoother shapes and fine geometric details.
Their 3D models are reconstructed from multi-view imagery.

\vspace{1mm}
\noindent\textbf{Instance count.}  
The number of object instances per image varies widely across datasets. LM-O, TUD-L, HB, and YCB-V contain at most one instance per object category.  
In contrast, ITODD contains on average 4 instances per object, with some scenes reaching up to 84 instances.  
T-LESS and IC-BIN contain 1.3 and 9 instances per object on average, respectively.

\vspace{1mm}
\noindent\textbf{Texture and appearance.}  
Objects in T-LESS and ITODD are texture-less, while those in LM-O, TUD-L, and HB are mostly uniformly colored.  
IC-BIN and YCB-V include richly textured objects, providing strong visual cues for correspondence.

\vspace{1mm}
\noindent\textbf{Scene complexity.}  
TUD-L includes single-object scenes captured under extreme lighting conditions.  
LM-O and HB feature highly cluttered environments with significant occlusion.  
IC-BIN, T-LESS, and ITODD contain multiple instances of the same or similar objects in close proximity, leading to heavy occlusion and symmetry-induced ambiguities.  
YCB-V features moderate levels of occlusion and clutter.

\subsection{Evaluation metrics}\label{sec:metrics}

We evaluate performance in terms of both accuracy and efficiency.  
For 6D localization, accuracy is measured using the Average Recall (AR) of three core metrics from the BOP Benchmark: VSD, MSSD, and MSPD.
For 6D detection, we report the Average Precision (AP) of MSSD and MSPD.
Each metric quantifies the discrepancy between the predicted and ground-truth object poses by comparing the corresponding 3D models after transformation.
Visible Surface Discrepancy (VSD) measures the difference between the depth maps rendered from the predicted and ground-truth poses, accounting for occlusions and visibility.
Maximum Symmetry-Aware Surface Distance (MSSD) computes the maximum Euclidean distance between corresponding 3D model points, taking into account all symmetrically equivalent ground-truth poses.
Maximum Symmetry-Aware Projection Distance (MSPD) measures the maximum 2D reprojection error between the predicted and ground-truth poses, also considering symmetric pose ambiguities.
In addition to accuracy, we report the run-time of each method to assess computational efficiency.

\subsection{Quantitative results}\label{sec:quantitative}
\renewcommand{\arraystretch}{0.9}
\begin{table*}[t]
    \centering
    \caption{
    Results on 6D localization on BOP datasets.
    We report the AR score on each of the seven core datasets of the BOP Benchmark, the mean AR and the mean per-image run-time across datasets . 
    Keys. 
    Training free: task-specific training free;
    Mean: Mean AR;
    *: Most updated results (and/or run-time) published on the BOP leaderboard;
    `-': not available.
    }
    \label{tab:loc6d}

    \vspace{-3mm}
    \resizebox{\textwidth}{!}{%
    \begin{tabular}{rlcc|ccccccc|cr}
        \toprule
        & \multirow{2}{*}{Method} & Training & Detection / & \multicolumn{7}{c|}{BOP Benchmark} & \multirow{2}{*}{Mean} &  \multirow{2}{*}{Time (s)} \\
        & & free & Segmentation & LM-O & T-LESS & TUD-L & IC-BIN & ITODD & HB & YCB-V & & \\
        \toprule  
        {\color{gray} \scriptsize 1} & MegaPose~\cite{labbe2022megapose} & & \multirow{3}{*}{ZeroPose} & 60.1 & 46.8 & 84.3 & 32.7 & 47.9 & 68.6 & 75.7 & 59.4 & 234.1 \\
        {\color{gray} \scriptsize 2} & ZeroPose~\cite{chen2024zeropose} & & & 58.3 & 49.6 & 72.5 & 44.9 & 51.5 & 64.0 & 79.0 & 60.0 & 85.9 \\      
        {\color{gray} \scriptsize 3} & SAM-6D~\cite{lin2024sam6d} & & & 63.5 & 43.0 & 80.2 & 51.8 & 48.4 & 69.1 & 79.2 & 62.2 & 5.5 \\
        \cmidrule{1-13}
        {\color{gray} \scriptsize 4} & ZeroPose*~\cite{chen2024zeropose} & & \multirow{8}{*}{CNOS} & 53.8 & 40.0	& 83.5 & 39.2 & 52.1 & 65.3 & 65.3 & 57.0 & 16.2\\
        {\color{gray} \scriptsize 5} & MegaPose*~\cite{labbe2022megapose} & & & 62.6 & 48.7 & 85.1 & 46.7 & 46.8 & 73.0 & 76.4 & 62.8 & 142.0 \\
        {\color{gray} \scriptsize 6} & MatchU~\cite{huang2024matchu} & & & 64.4 & 52.7 & 89.8 & 44.2 & - & - & 72.6 & - & -  \\
        {\color{gray} \scriptsize 7} & SAM-6D~\cite{lin2024sam6d} & & & 65.1 & 47.9 & 82.5 & 49.7 & 56.2 & 73.8 & 81.5 & 65.3 & 1.3 \\
        {\color{gray} \scriptsize 8} & GenFlow*~\cite{moon2024genflow} & & & 63.5 & 52.1 & 86.2 & 53.4 & 55.4 & 77.9 & 83.3 & 67.4 & 34.6 \\
        {\color{gray} \scriptsize 9} & GigaPose+GenFlow+kabsch*~\cite{moon2024genflow} & & & 67.8 & 55.6 & 81.1 & 56.3 & 57.5 & 79.1 & 82.5 & 68.6 & 11.1 \\
        {\color{gray} \scriptsize 10} & FreeZe~\cite{caraffa2024freeze} & \cmark & & 69.0 & 52.0 & 93.6 & 49.9 & 56.1 & 79.0 & 85.3 & 69.3 & 13.5 \\
        \rowcolor{myazure} {\color{gray} \scriptsize 11} & \acronym  & \cmark & & 69.4 & 56.3 & 93.5 & 54.5 & 60.0 & 76.9 & 87.4 & 71.1 & 1.5 \\
        \cmidrule{1-13}
        {\color{gray} \scriptsize 12} & SAM-6D~\cite{lin2024sam6d} & & \multirow{4}{*}{SAM-6D} & 69.9 & 51.5 & 90.4 & 58.8 & 60.2 & 77.6 & 84.5 & 70.4 & 4.4 \\
        {\color{gray} \scriptsize 13} & FoundationPose*~\cite{wen2024foundationpose} & & & 75.6 & 64.6 & 92.3 & 50.8 & 58.0 & 83.5 & 88.9 & 73.4 & 29.3 \\
        {\color{gray} \scriptsize 14} & FreeZe~\cite{caraffa2024freeze} & \cmark & & 71.6 & 53.1 & 94.9 & 54.5 & 58.6 & 79.6 & 84.0 & 70.9 & 11.5 \\
        \rowcolor{myazure} {\color{gray} \scriptsize 15} & \acronym  & \cmark & & 73.3 & 59.7 & 95.2 & 60.9 & 63.3 & 80.1 & 87.8 & 74.3 & 1.4 \\
        \cmidrule{1-13}
        \rowcolor{myazure} {\color{gray} \scriptsize 16} & \acronym  & \cmark & NIDS & 68.4 & 61.5 & 94.5 & 59.0 & 56.7 & 84.3 & 89.3 & 73.4 & 1.3 \\
        \midrule 
        \rowcolor{myazure} {\color{gray} \scriptsize 17} & \acronym  & \cmark & MUSE & 73.1 & 61.6 & 95.5 & 61.6 & 63.6 & 78.2 & 89.3 & 74.7 & 1.5 \\
        \midrule 
        \rowcolor{myazure} {\color{gray} \scriptsize 18} & \acronym  & \cmark & CNOS, SAM-6D, & 75.9 & 72.9 & 97.2 & 66.3 & 71.2 & 86.0 & 91.3 & 80.1 & 4.4 \\
        \rowcolor{myazure} {\color{gray} \scriptsize 19} & \acronym-Accurate  & \cmark & NIDS, MUSE & 77.1 & 75.5 & 97.6 & 69.7 & 74.2 & 89.2 & 91.5 & 82.1 & 24.8 \\
        \bottomrule
    \end{tabular}
}
\end{table*}

\renewcommand{\arraystretch}{0.9}
\begin{table*}[t]
    \centering
    \caption{
    Results on 6D detection on BOP datasets, where no prior information about the number of instances is available.
    We report the AP score on each of the seven core datasets of the BOP Benchmark, the mean AP and the mean per-image run-time across datasets . 
    Keys. 
    Training free: task-specific training free;
    Prior: type of instance localization prior;
    Refin.: pose refinement; 
    Mean: Mean AP;
    *: Most updated results (and/or run-time) published on the BOP leaderboard;
    C: CNOS;
    S: SAM-6D;
    N: NIDS;
    M: MUSE.
    }
    \label{tab:det6d}

    \vspace{-3mm}
    \resizebox{\textwidth}{!}{%
    \begin{tabular}{rlcc|ccccccc|cr}
        \toprule
        & \multirow{2}{*}{Method}  & Training & Detection / & \multicolumn{7}{c|}{BOP Benchmark} & \multirow{2}{*}{Mean} &  \multirow{2}{*}{Time (s)} \\
        & & free & Segmentation & LM-O & T-LESS & TUD-L & IC-BIN & ITODD & HB & YCB-V & & \\
        \toprule  
        {\color{gray} \scriptsize 1} & GigaPose+GenFlow+kabsch*~\cite{moon2024genflow} & & & 23.7 & 52.6 & 24.6 & 29.2 & 52.5 & 54.7 & 52.4 & 41.4 & 16.6 \\
        {\color{gray} \scriptsize 2} & GigaPose+GenFlow*~\cite{moon2024genflow} & & & 57.8 & 46.7 & 71.5 & 40.2 & 44.7 & 67.6 & 68.2 &  56.7 & 4.5 \\
        \rowcolor{myazure} {\color{gray} \scriptsize 3} & \acronym  & \cmark & \multirow{-3}{*}{CNOS}  & 71.2 & 56.4 & 89.4 & 44.1 & 58.3 & 72.2 & 85.8 & 68.2 & 4.0 \\
        \cmidrule{1-13}
        \rowcolor{myazure} {\color{gray} \scriptsize 4} & \acronym-Accurate  & \cmark & C+S+N+M & 79.7 & 75.1 & 99.1 & 69.6 & 76.9 & 85.3 & 90.5 & 82.3 & 37.3 \\
        \bottomrule
    \end{tabular}
}

\end{table*}

\noindent\textbf{6D Localization.}  
We report results for 6D localization in Tab.~\ref{tab:loc6d}.  
\acronym is compared against published works on the seven core datasets of the BOP Benchmark. 
Competitors are grouped by the detector used:  
ZeroPose~\cite{chen2024zeropose} (Rows 1–3),  
CNOS~\cite{nguyen2023cnos} (Rows 4–11),  
SAM-6D~\cite{lin2024sam6d} (Rows 12–15),  
NIDS~\cite{lu2024adapting} (Row 16)
and MUSE~\cite{muse_bop2024} (Row 17).
Rows 18–19 report the performance of \acronym when using an ensemble of segmentation models, which is one of the key novelties introduced in this work.  
All methods are training-based, with the exception of FreeZe~\cite{caraffa2024freeze} and \acronym; all methods operate in the RGBD setting.
\acronym consistently outperforms all other methods in terms of both accuracy (Column ‘Mean’) and efficiency (Column ‘Time’).
Among methods based on CNOS (Rows 4–11), \acronym (Row 11) achieves 71.1 AR in 1.5 seconds, outperforming all competitors.  
Compared to FreeZe~\cite{caraffa2024freeze} (Row 10), \acronym improves accuracy by +1.8 AR while reducing runtime by nearly 9×.
It also surpasses GenFlow-based variants (Rows 8–9), which are slower and less accurate despite relying on extensive task-specific training.
The most efficient baseline, SAM-6D (Row 7), runs slightly faster at 1.3 seconds, but falls short by 5.8 AR, showing that \acronym offers a significantly better speed–accuracy trade-off.
Among methods using SAM-6D (Rows 12–15), \acronym (Row 15) achieves 74.3 AR, outperforming all prior works.  
It improves upon FoundationPose~\cite{wen2024foundationpose} (Row 13), the strongest training-based competitor, by +0.9 AR while running more than 20× faster (1.4 s vs. 29.3 s).  
Compared to FreeZe (Row 14), \acronym gains +3.4 AR with an 8× speed-up, demonstrating how sparse target sampling and feature-aware scoring boost both efficiency and alignment accuracy.
When using the NIDS detector (Row 16), \acronym reaches 73.4 AR, while with MUSE (Row 17), it achieves 74.7 AR, establishing new state-of-the-art results using a single segmentation model.
These results show that \acronym is robust to different segmentation priors.
With an ensemble of detectors (Row 18), \acronym achieves 80.1 AR in 4.4 seconds, outperforming all methods based on single segmentation models.
Despite processing masks from four independent detectors, \acronym remains significantly faster than most training-based competitors.
Compared to FoundationPose (Row 13), which achieves 73.4 AR in 29.3 seconds using SAM-6D detections, \acronym gains +6.7 AR while running over 6 times faster.
Compared to FreeZe (Row 14), \acronym improves by +9.2 AR with a nearly 3× speed-up.
Even with the larger number of input masks, \acronym matches the runtime of the SAM-6D baseline (Row 12), which runs in 4.4 seconds but is significantly less accurate, trailing by 9.7 AR.
These results demonstrate the scalability and efficiency of \acronym, which remains robust and fast even when integrating multiple segmentation sources.
The \acronym-Accurate configuration (Row 19), winner of the BOP Challenge 2024~\cite{nguyen2025bop} (entry FreeZeV2.1 in the leaderboard\footnote{\url{https://bop.felk.cvut.cz/method_info/905/}, accessed: 1st June 2025.}), pushes performance to 82.1 AR.  
In addition to all improvements introduced in \acronym, it integrates Symmetry-Aware Refinement (SAR)~\cite{caraffa2024freeze}, increases the number of processed masks up to $M = 2N$, and improves scoring by comparing visual features of the input image and the rendered pose.  
Although slightly slower at 24.8 seconds per image, it remains faster than many prior methods, including FoundationPose (29.3 s) and GenFlow (34.6 s), while setting a new state-of-the-art in accuracy.
These results confirm that \acronym’s feature-aware scoring strategy enables effective integration of redundant and complementary detections across models, without requiring any heuristic fusion. 
Across all segmentation settings, \acronym consistently achieves high accuracy and fast inference, validating the generalization strength of its training-free design and its suitability for real-time, zero-shot 6D localization in open-world scenarios.

\vspace{1mm}
\noindent\textbf{6D Detection.}  
Tab.~\ref{tab:det6d} reports results for 6D detection, where both object identity and pose must be inferred without prior knowledge of the number of instances per image.
We compare \acronym against two variants of GenFlow~\cite{moon2024genflow}, the only published methods with leaderboard entries for this task. All compared methods in this setting use the CNOS detector.
\acronym (Row 3) achieves 68.2 AP in 4.0 seconds per image, significantly outperforming both GigaPose+GenFlow (Row 2, 56.7 AP) and its Kabsch variant (Row 1, 41.4 AP).
It is also the fastest among all methods, offering a substantial +11.5 AP accuracy gain over the strongest competitor at comparable or lower latency.
This confirms that our training-free pipeline generalizes well even in the more challenging 6D detection scenario.
Although the runtime increases to 37.3 seconds per image, the improvement in accuracy justifies the additional cost, especially considering that no task-specific training is required.
The longer inference time in detection compared to localization is due to the need to process a larger number of candidate masks, since the number of instances is unknown.
From the perspective of \acronym, this is the only difference between the two tasks.
It is handled through the hyperparameter $M$, which controls how many masks are processed.
The rest of the pipeline remains unchanged, showing that \acronym is a unified, efficient, and scalable solution for both tasks.

\subsection{Qualitative results}\label{sec:qualitative}

\begin{figure*}[t]
\centering
    \begin{tabular}{@{}c@{\,}c@{\,}c@{\,}c}
    \raggedright
        \begin{overpic}[width=0.24\textwidth]
        {figures/qualitatives/001098_input_8.png}
            \put(-7, 27){\rotatebox{90}{Input}}
        \end{overpic} &
        \begin{overpic}[width=0.24\textwidth]
        {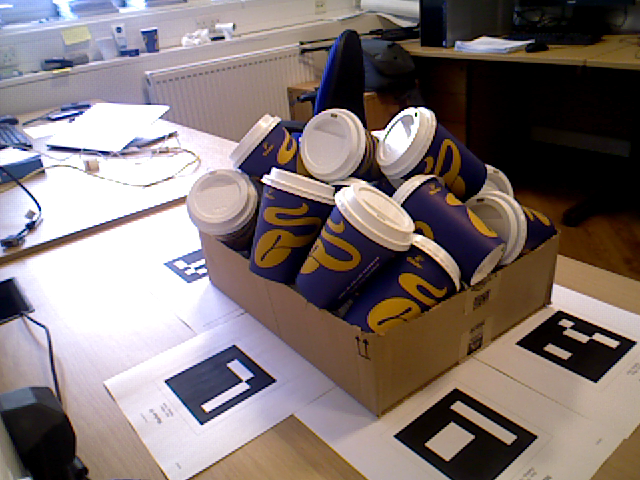}
        \end{overpic} &
        \begin{overpic}[width=0.24\textwidth]
        {figures/qualitatives/001709_input_5.png}
        \end{overpic} &
        \begin{overpic}[width=0.24\textwidth]
        {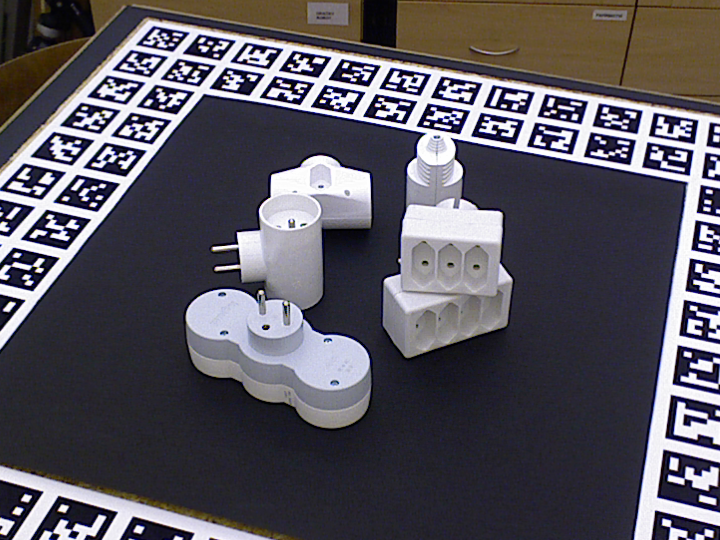}
        \end{overpic}

        \\

        \begin{overpic}[width=0.24\textwidth]
        {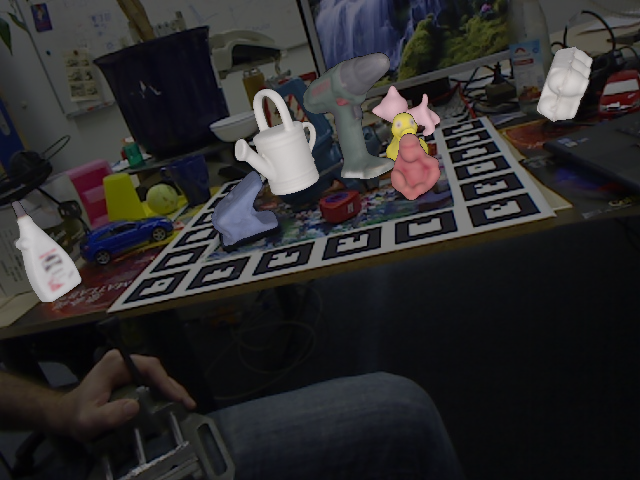}
            \put(-7, 27){\rotatebox{90}{Output}}
            \put(-2.5, 3){\colorbox{white}{(a)}}
        \end{overpic} &
        \begin{overpic}[width=0.24\textwidth]
        {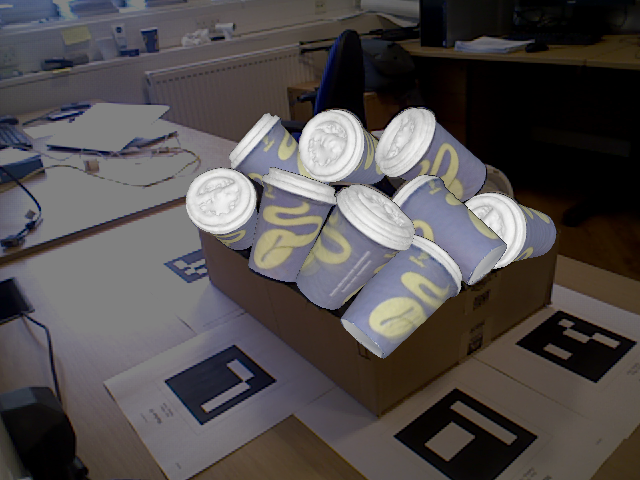}
            \put(-2.5, 3){\colorbox{white}{(b)}}
        \end{overpic} &
        \begin{overpic}[width=0.24\textwidth]
        {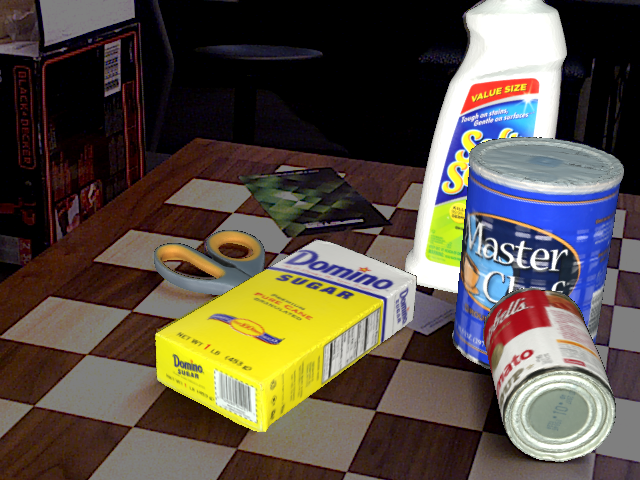}
            \put(-2.5, 3){\colorbox{white}{(c)}}
        \end{overpic} &
        \begin{overpic}[width=0.24\textwidth]
        {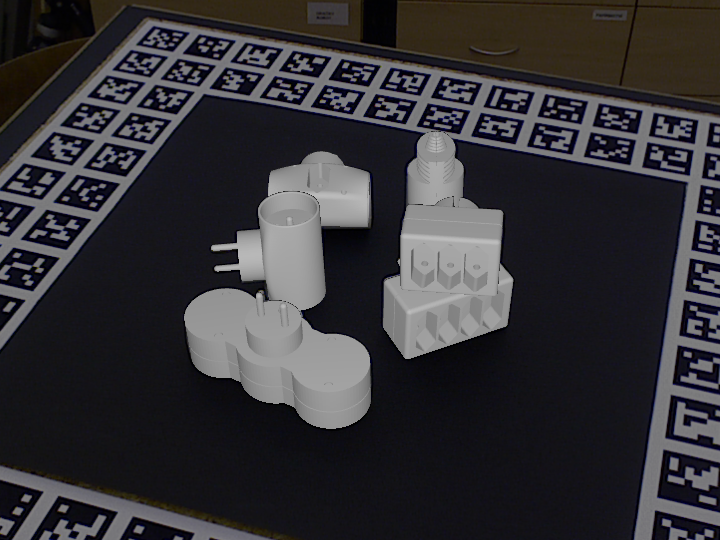}
            \put(-2.5, 3){\colorbox{white}{(d)}}
        \end{overpic} 
        
    \end{tabular}
    \vspace{-3mm}
    \caption{
    Qualitative results on several sample images from the BOP benchmark.
    The top row shows input images, while the bottom row shows \acronym's predictions, obtained by overlaying the corresponding 3D models transformed according to the predicted 6D poses. Each image presents unique challenges: severe occlusions (a, b, c, d), 
    multiple instances of the same object in close proximity (b),
    geometric object symmetry (b, c),
    common everyday household items (a, b, c),
    industrial objects with complex shapes (d),
    objects with uniform color (a),
    objects with detailed texture (b, c),
    and texture-less objects (d).
    }
    \label{fig:qualitative}
\end{figure*}

Fig.~\ref{fig:qualitative} illustrates qualitative results on representative samples from the BOP Benchmark, highlighting the generalizability of \acronym across a wide range of object types and scene conditions.
The top row shows the input RGB images, while the bottom row visualizes the estimated poses by overlaying the 3D object models rendered using the predicted transformations.
Each scene presents distinct challenges.
Scene (a) features a tabletop environment with multiple everyday objects, several of which are partially occluded or overlapping.
Despite visual clutter and the presence of distractors such as a human hand and background markers, \acronym produces consistent predictions. 
The accurate alignment of texture-less objects (e.g., the white watering can) and textured ones (e.g., the yellow duck) demonstrates the method's ability to leverage both geometric and visual cues, even under moderate occlusion.
Scene (b) contains multiple instances of a symmetric object arranged tightly inside a cardboard box.
The repetitive appearance and rotational symmetry make this a challenging case.
\acronym is able to correctly estimate individual poses for all visible instances without confusion or collapse, showing robustness to instance-level ambiguity and strong prior-free symmetry handling.
This suggests that geometric reasoning plays a key role in distinguishing similar poses even in tightly packed settings.
Scene (c) includes a set of everyday objects with highly discriminative textures (e.g., branded packaging) as well as symmetric shapes (e.g., cans).
Notably, objects with similar geometry but different textures are correctly resolved, indicating that \acronym effectively integrates appearance cues from vision foundation models to disambiguate poses in cluttered, real-world contexts.
(d) shows texture-less industrial components arranged on a black planar surface. These objects lack any distinctive texture and exhibit substantial symmetry and surface uniformity.
\acronym accurately estimates their 6D poses, suggesting that the geometric features extracted from the foundation model are sufficient to guide alignment in the absence of visual texture.
Overall, these results demonstrate that \acronym performs reliably across a broad spectrum of challenges including occlusion, symmetry, instance duplication, self-similarity, and texture scarcity.
\acronym achieves strong generalization across both household and industrial objects by integrating geometric and visual information, without requiring fine-tuning on task-specific data.

\subsection{Ablation Study}\label{sec:ablation}
\begin{table}[t]
    \centering
    \tabcolsep 3pt
    \caption{
    Ablation on pose scoring and matching configuration in 6D localization.
    We vary the score used whitin RANSAC, the weights $\alpha$, $\beta$, and $\gamma$ in Eq.~\eqref{eq:final_score} used to combine coarse score, fine score, and ICP score, and the number of nearest neighbors $k$ used for feature matching.
    }
    \label{tab:topk_results}
    \vspace{-3mm}
    \resizebox{\columnwidth}{!}{%
    \begin{tabular}{r@{\hskip 1mm}l|ccccc|c}
        \toprule
        & Configuration & LM-O & T-LESS & TUD-L & IC-BIN & YCB-V & Mean \\
        \toprule
        \multicolumn{7}{l}{\textit{RANSAC score ablation (standard inlier ratio)}} \\
        {\color{gray} \scriptsize 1} & $k{=}10$, $\alpha{=}0$, $\beta{=}0$, $\gamma{=}1$ & 70.5 & 57.8 & 91.5 & 55.2 & 83.9 & 71.9 \\
        {\color{gray} \scriptsize 2} & $k{=}10$, $\alpha{=}1$, $\beta{=}1$, $\gamma{=}1$ & 72.0 & 59.0 & 94.8 & 55.4 & 84.7 & 73.2 \\
        \midrule
        \multicolumn{7}{l}{\textit{Final score weights ablation (feature-aware)}} \\
        {\color{gray} \scriptsize 3} & $k{=}10$, $\alpha{=}1$, $\beta{=}0$, $\gamma{=}0$ & 73.1 & 59.6 & 94.0 & 60.0 & 85.6 & 74.5 \\
        {\color{gray} \scriptsize 4} & $k{=}10$, $\alpha{=}0$, $\beta{=}1$, $\gamma{=}0$ & 73.1 & 59.7 & 94.1 & 60.4 & 85.9 & 74.6 \\
        {\color{gray} \scriptsize 5} & $k{=}10$, $\alpha{=}1$, $\beta{=}1$, $\gamma{=}0$ & 73.2 & 59.5 & 94.3 & 60.4 & 85.9 & 74.7 \\
        {\color{gray} \scriptsize 6} & $k{=}10$, $\alpha{=}0$, $\beta{=}0$, $\gamma{=}1$ & 71.2 & 58.5 & 91.5 & 60.3 & 86.9 & 73.7 \\
        \midrule
        \multicolumn{7}{l}{\textit{Top-K matching ablation}} \\
        {\color{gray} \scriptsize 7} & $k{=}1$, $\alpha{=}1$, $\beta{=}1$, $\gamma{=}1$ & 72.8 & 59.5 & 94.6 & 59.4 & 87.8 & 74.7 \\
        \midrule
        {\color{gray} \scriptsize 8} & \textbf{$k{=}10$, $\alpha{=}1$, $\beta{=}1$, $\gamma{=}1$ }(ours) & \textbf{73.3} & \textbf{59.7} & \textbf{95.2} & \textbf{60.9} & \textbf{87.8} & \textbf{75.4} \\
        \bottomrule
    \end{tabular}    
    }

\end{table}

\begin{table}[t]
    \centering
    \tabcolsep 3pt
    \caption{
    Ablation on the number $M$ of target candidate masks considered for 6D localization.
    }
    \label{tab:num_masks}
    \vspace{-3mm}
    \resizebox{\columnwidth}{!}{%
    \begin{tabular}{r@{\hskip 1mm}l|ccccc|cc}
        \toprule
        & Segmentation masks $M$ & LM-O & T-LESS & TUD-L & IC-BIN & YCB-V & Mean & Time \\
        \toprule    
        {\color{gray} \scriptsize 1} & $M=N$ & 71.6 & 58.9 & 92.1 & 59.7 & 86.4 & 73.7 & 1.2 \\
        {\color{gray} \scriptsize 2} & $M=N+1$ (ours) & 73.3 & 59.7 & 95.2 & 60.9 & 87.8 & 75.4 & 1.5 \\
        {\color{gray} \scriptsize 3} & $M=N+2$ & 73.7 & 60.1 & 95.7 & 60.8 & 87.8 & 75.6 & 1.7 \\
        \bottomrule
    \end{tabular}    
    }

\end{table}

Tab.~\ref{tab:topk_results} presents a detailed ablation study on two key design choices in our method: the feature-aware scoring function used to rank pose hypotheses, and the correspondence strategy adopted for feature matching.
Specifically, we evaluate the individual contribution of each component in the final scoring function defined in Eq.~\eqref{eq:final_score}, as well as the impact of using top-$k$ versus single-nearest-neighbor matching for establishing correspondences.
We conduct the ablation on five representative datasets from the BOP Benchmark, namely LM-O~\cite{brachmann2014learning}, T-LESS~\cite{hodan2017tless}, TUD-L~\cite{tudl}, IC-BIN~\cite{doumanoglou2016recovering}, and YCB-V~\cite{xiang2018posecnn}, covering a diverse range of object types and scene conditions.

\vspace{0.5mm}
\noindent\textbf{Effect of feature-aware scoring.}  
We evaluate our novel feature-aware score, defined in Eq.~\eqref{eq:score_feat}, at two distinct stages: (i) within RANSAC, where it guides the selection of the best coarse pose for a given candidate mask, and (ii) during final pose ranking across masks.  
To assess its contribution within RANSAC, we replace $S_\text{feat}^\text{coarse}$ with a standard inlier ratio as in classical RANSAC.  
For final pose ranking, we either use only $S_\text{ICP}$ (Row~1, 71.9 AR), or an alternative final score that replaces $S_\text{feat}^\text{coarse}$ and $S_\text{feat}^\text{fine}$ with standard inlier ratios over $\mathcal{P}_T^\text{sparse}$ (Row~2, 73.2 AR).  
While Row~2 yields a 1.3 AR gain over Row~1, it underperforms when using our feature-aware score in both stages (Row~8, 75.4 AR).  
These results confirm that our feature-aware score $S_\text{feat}$ plays a critical role not only in guiding RANSAC toward better coarse poses, but also in accurately ranking the final set of candidate poses across masks.

\vspace{0.5mm}
\noindent\textbf{Component-wise analysis of the scoring function.}  
Rows~3--6 isolate the contribution of each individual score component used in Eq.~\eqref{eq:final_score}.  
Using only $S_\text{feat}^\text{coarse}$ (Row~3) or only $S_\text{feat}^\text{fine}$ (Row~4) results in 74.5 and 74.6 AR, respectively, while combining both (Row~5) increases accuracy to 74.7 AR, suggesting that both accurate coarse and fine pose are determinant.  
In contrast, using only the ICP alignment score $S_\text{ICP}$ (Row~6) leads to lower accuracy (73.7 AR), though it still outperforms the baseline in Row~1, thanks to the use of $S_\text{feat}^\text{coarse}$ in RANSAC.
These results show that our feature-aware similarity terms are more effective than geometric alignment alone.
Notably, $S_\text{feat}^\text{fine}$ performs slightly better than $S_\text{feat}^\text{coarse}$ (Row~4 vs Row~3), and combining all three terms in Row~8 yields the best overall result with 75.4 AR.

\vspace{0.5mm}
\noindent\textbf{Effect of top-$k$ matching.}  
Row~7 evaluates single-nearest-neighbor matching ($k{=}1$), yielding 74.7 AR.  
Using top-$k$ matching with $k{=}10$ (Row~8) increases accuracy to 75.4 AR with a +0.7 improvement.  
This confirms that considering multiple correspondences increases robustness to feature noise.

\vspace{0.5mm}
\noindent\textbf{Candidate mask selection.}  
Tab.~\ref{tab:num_masks} analyzes the impact of varying the number $M$ of segmentation masks considered.
When $M = N$ (Row~1), the final pose ranking is bypassed, as only the $N$ most confident masks from the segmentation model are retained. This yields 73.7 AR with an inference time of 1.2s.  
Increasing to $M = N+1$ (Row~2) introduces our feature-aware score into the mask selection process, improving accuracy to 75.4 AR (+1.7) with only a slight increase in time (1.5s).  
Using $M = N+2$ (Row~3) further boosts AR to 75.6, though inference time increases to 1.7s.  
These results highlight the effectiveness of our scoring function in identifying the correct pose and filtering out suboptimal hypotheses.

\vspace{0.5mm}
\noindent\textbf{Summary.}  
The ablation confirms that both the feature-aware score and the top-$k$ strategy are essential for accurate pose estimation.  
Feature similarity provides stronger alignment cues than geometric consistency alone, and using multiple matches improves robustness without task-specific tuning. Moreover, our score effectively supports cross-mask pose ranking.

\section{Conclusions}\label{sec:conclusions}

We presented \acronym, a zero-shot 6D pose estimation method that leverages frozen foundation models without relying on large-scale training on task-specific synthetic data.
Compared to its predecessor FreeZe, \acronym introduces key innovations that significantly enhance both accuracy and efficiency.
Specifically, \acronym employs a sparse-to-dense feature matching strategy to significantly speed-up inference, a feature-aware scoring mechanism to enhance the selection and ranking of pose hypothesis, and integrates ensembles of zero-shot segmentation models to yield more reliable localization priors.
We demonstrated the effectiveness of \acronym through comprehensive evaluations on the BOP Benchmark, the gold standard for 6D pose estimation, where it consistently outperformed all publicly available methods in both the 6D localization and 6D detection tasks.
Notably, \acronym was awarded \emph{Best Overall Method} at the BOP Challenge 2024 in both categories, highlighting its competitiveness even against training-based approaches.

While \acronym significantly improves runtime over previous zero-shot methods, it still falls short of the real-time requirements demanded by many industrial applications.
Additionally, its performance is strongly influenced by the quality of the zero-shot segmentation used to localize the object of interest within the input image.
Although using ensembles of segmentation models improves accuracy, it comes at the cost of increased computational overhead.
Addressing these limitations offers promising directions for future work.

\bibliographystyle{IEEEtran}
\bibliography{main}

\end{document}